%% file: arxiv.tex
\documentclass[10pt,twocolumn,letterpaper]{article}

\usepackage[pagenumbers]{cvpr} %

\usepackage{graphicx}
\usepackage{amsmath}
\usepackage{amssymb}
\usepackage{booktabs}
\usepackage{xcolor}
\usepackage{multirow}
\usepackage{makecell}
\usepackage{overpic}

\makeatletter
    \newcommand{\vast}{\bBigg@{3}}
    \newcommand{\Vast}{\bBigg@{3.5}}
    \newcommand{\vastt}{\bBigg@{4}}
    \newcommand{\Vastt}{\bBigg@{4.5}}
\makeatother

\definecolor{darkgreen}{RGB}{30,150,30}
\definecolor{darkblue}{RGB}{0,0,127}
\definecolor{darkyellow}{RGB}{171,133,0}
\definecolor{darkred}{RGB}{180,20,20}
\definecolor{darkmagenta}{RGB}{200,0,200}
\definecolor{darkcyan}{RGB}{0,127,127}

\newif\ifdrafting 
\draftingtrue %

\ifdrafting
  \newcommand{\GL} [1] {\textcolor{darkgreen}{[GL: #1]}}
  \newcommand{\VJ} [1] {\textcolor{darkblue}{[VJ: #1]}}
  \newcommand{\ls} [1] {\textcolor{darkmagenta}{[LS: #1]}}
  \newcommand{\ds} [1] {\textcolor{red}{[DS: #1]}}
  \newcommand{\TODO} [1] {{\color{darkcyan}{\bf [TODO: #1]}}}
  
\else
  \newcommand{\GL} [1] {}
  \newcommand{\VJ} [1] {}
  \newcommand{\ls} [1] {}
  \newcommand{\ds} [1] {}  
  \newcommand{\TODO} [1] {}
  
\fi

\usepackage{tikz}
\usepackage{pifont}

\newcommand{\p}[1]{OOAL}
\newcommand{\s}[1]{PartSelect}

\usepackage[pagebackref,breaklinks,colorlinks]{hyperref}

\usepackage[capitalize]{cleveref}
\crefname{section}{Sec.}{Secs.}
\Crefname{section}{Section}{Sections}
\crefname{table}{Tab.}{Tabs.}
\Crefname{table}{Table}{Tables}

\def\cvprPaperID{477} %
\def\confName{CVPR}
\def\confYear{2024}

\begin{document}

\title{One-Shot Open Affordance Learning with Foundation Models}
\author{Gen Li\textsuperscript{1}\qquad
Deqing Sun\textsuperscript{2}\qquad
Laura Sevilla-Lara\textsuperscript{1} \qquad Varun Jampani\textsuperscript{3}\\ \\
\textsuperscript{1}University of Edinburgh\quad
\textsuperscript{2}Google Research\quad
\textsuperscript{3}Stability AI\\
}

\twocolumn[{%
\renewcommand\twocolumn[1][]{#1}%
\maketitle
}]

\begin{abstract}
We introduce One-shot Open Affordance Learning (OOAL), where a model is trained with just one example per base object category, but is expected to identify novel objects and affordances.
While vision-language models excel at recognizing novel objects and scenes, they often struggle to understand finer levels of granularity such as affordances.
To handle this issue, we conduct a comprehensive analysis of existing foundation models, to explore their inherent understanding of affordances and assess the potential for data-limited affordance learning.
We then propose a vision-language framework with simple and effective designs that boost the alignment between visual features and affordance text embeddings. 
Experiments on two affordance segmentation benchmarks 
show that the proposed method outperforms state-of-the-art models with less than 1\% of the full training data, and exhibits reasonable generalization capability on unseen objects and affordances. 
\end{abstract}

\input{1_introduction}

\input{2_related_work}

\input{3_proposed_method}

\input{4_experiments}

\input{5_conclusion}

{\small
\bibliographystyle{ieee_fullname}
\bibliography{egbib}
}
\clearpage

\appendix
\renewcommand\thesection{\Alph{section}}
\pagestyle{empty}
\thispagestyle{empty}

\section{Dataset Details}
\label{sec:exp_detail}

\input{dataset_tab}
To evaluate the model's generalization ability in the challenging One-shot Open Affordance Learning (OOAL) setting, datasets with a large number of object categories are required.
In addition, at least two object categories are needed for each affordance so that the model can be trained on one object and tested on the other.
After an investigation of existing affordance datasets, we find only two datasets, AGD20K \cite{ag_from_exocentric_imgs} and UMD \cite{tool_parts_iff}, that fulfill the prerequisites and can be used to evaluate the affordance segmentation task.
Specific affordance and object categories of these two datasets are shown in \cref{class_dataset}.
For the unseen split, we display the object category division in \cref{unseen_dataset}.
The model is trained on base object classes, and evaluated on novel objects categories.

\begin{figure}[t]
  \centering
   \includegraphics[width=1.0\linewidth]{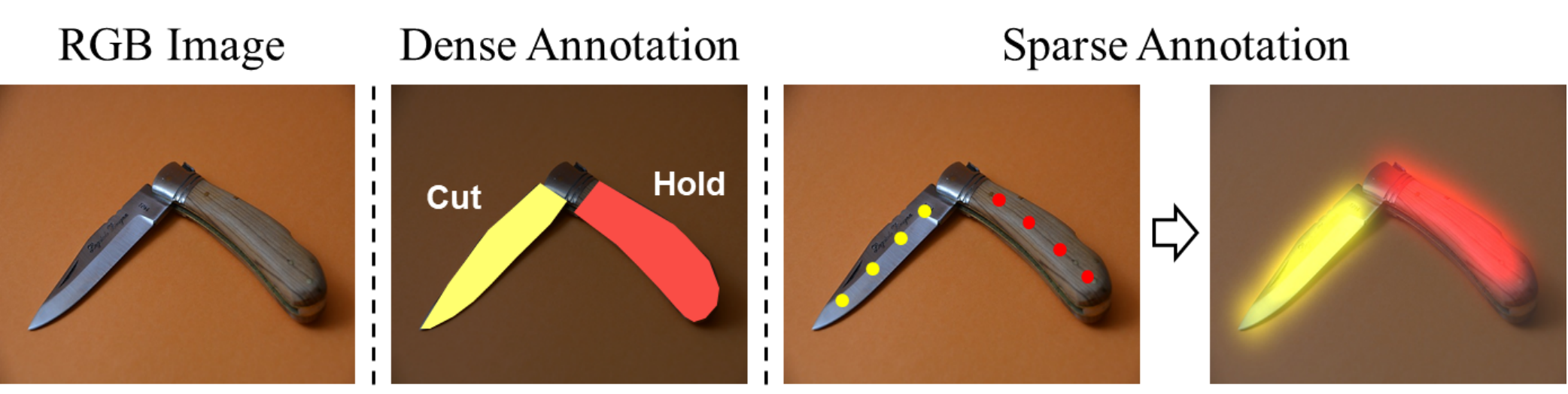}
   \caption{Different affordance annotation schemes. Dense affordance annotation is labeled as binary masks. Sparse affordance annotation is first labeled as keypoints, and then a gaussian kernel is performed over each point to produce pixel-wise ground truth.}
   \label{fig:task}
\end{figure}

Moreover, it is worth noting that annotations in AGD20K and UMD are of different types. 
UMD uses pixel-level dense binary maps, while the ground truth of AGD20K consist of sparse keypoints within the affordance areas, and a gaussian distribution is then applied on each point to generate dense annotation.
The difference of dense and sparse affordance annotation is highlighted in \cref{fig:task}.

\section{Ablation Study on Hyperparameters}
\label{sec:supp_exp}
The proposed framework involves three primary hyperparameters, \ie, the number of learnable text tokens $p$, vision encoder fusion layers $j$, decoder transformer layers $t$.
We conduct ablation studies individually to explore the impact of these hyperparameters, as detailed in \cref{tab:hyper_coop}, \cref{tab:hyper_mlff}, and \cref{tab:hyper_decoder_layer}.
Notably, increasing the number of learnable text tokens up to 8 showcases a gradual improvement in performance within the seen setting, but leads to fluctuating results in the unseen setting, indicating its susceptibility to generalization when confronted with unseen objects.
In terms of the fusion layers, the fusion of the last two layers demonstrates an obvious performance gain compared to the single-layer counterpart, and integrating the last three layers yields the best results.
Lastly, we note that the transformer decoder can effectively improve performance in both seen and unseen setting, and a two-layer transformer decoder produces the most optimal results.

\begin{table}[!t]
\centering
\small
\begin{tabular}{ccccccc}
\toprule
\multirow{2.3}{*}{$p$} & \multicolumn{3}{c}{Seen} & \multicolumn{3}{c}{Unseen} \\ \cmidrule(lr){2-4} \cmidrule(lr){5-7}  
                        & KLD$\downarrow$    & SIM$\uparrow$    & NSS$\uparrow$    & KLD$\downarrow$     & SIM$\uparrow$     & NSS$\uparrow$    \\ \midrule
2                     &     0.774   &  0.568      &  1.710      &     1.119    &    0.457     &  1.434      \\
4                   &  0.765      &  0.573     &  1.714      &   1.102      &  \textbf{0.469}       & 1.449    \\
6               &    0.760    &  0.572     &  1.726      &     1.162    &  0.440       & 1.383       \\
8                &    \textbf{0.740} & 0.577       &   \textbf{1.745}     &  \textbf{1.070}       &    0.461     & \textbf{1.503}\\
10                &  0.768    & \textbf{0.581}     & 1.726       &  1.111       &    0.460     & 1.463
\\ \bottomrule
\end{tabular}
\caption{Ablation study on the number of learnable token $p$ in text prompt learning. 
}
\label{tab:hyper_coop}
\end{table}

\begin{table}[!t]
\centering
\small
\begin{tabular}{ccccccc}
\toprule
\multirow{2.3}{*}{$j$} & \multicolumn{3}{c}{Seen} & \multicolumn{3}{c}{Unseen} \\ \cmidrule(lr){2-4} \cmidrule(lr){5-7}  
                        & KLD$\downarrow$    & SIM$\uparrow$    & NSS$\uparrow$    & KLD$\downarrow$     & SIM$\uparrow$     & NSS$\uparrow$    \\ \midrule
1                     & 1.060       & 0.455        &  1.422      &     1.338    &    0.390     &  1.302      \\
2                   & 0.748       & 0.576      &  \textbf{1.756}      &   1.105      &  0.456       & 1.452    \\
3               &  \textbf{0.740}      & 0.577      &  1.745      &     \textbf{1.070}    &  \textbf{0.461}       & \textbf{1.503}       \\
4                & 0.762      & \textbf{0.579}       & 1.713       &  1.129       &    0.453     & 1.401 \\ \bottomrule
\end{tabular}
\caption{Ablation study on the number of fusion layers $j$ in multi-layer feature fusion. 
}
\label{tab:hyper_mlff}
\end{table}

\begin{table}[!t]
\centering
\small
\begin{tabular}{ccccccc}
\toprule
\multirow{2.3}{*}{$t$} & \multicolumn{3}{c}{Seen} & \multicolumn{3}{c}{Unseen} \\ \cmidrule(lr){2-4} \cmidrule(lr){5-7}  
                        & KLD$\downarrow$    & SIM$\uparrow$    & NSS$\uparrow$    & KLD$\downarrow$     & SIM$\uparrow$     & NSS$\uparrow$    \\ \midrule
0                     &    0.846    &  0.537      & 1.622       &     1.115    &    0.447     &  1.440      \\
1                   &    0.753    &  0.574     & 1.737       & 1.094       & 0.449       & \textbf{1.492}   \\
2               &  \textbf{0.740}      &   \textbf{0.577}    & \textbf{1.745}       &     \textbf{1.067}    &  \textbf{0.465}       & \textbf{1.492}       \\
3                & 0.746     & 0.575       & 1.738       &  1.110       &    0.458     & 1.456 \\ \bottomrule
\end{tabular}
\caption{Ablation study on the number of transformer decoder layers $t$. 
}
\label{tab:hyper_decoder_layer}
\end{table}

\section{Additional Visualizations}
\label{sec:supp_vis}

\subsection{Visualization of CLS-guided mask}
\begin{figure}[t]
  \centering
   \includegraphics[width=.95\linewidth]{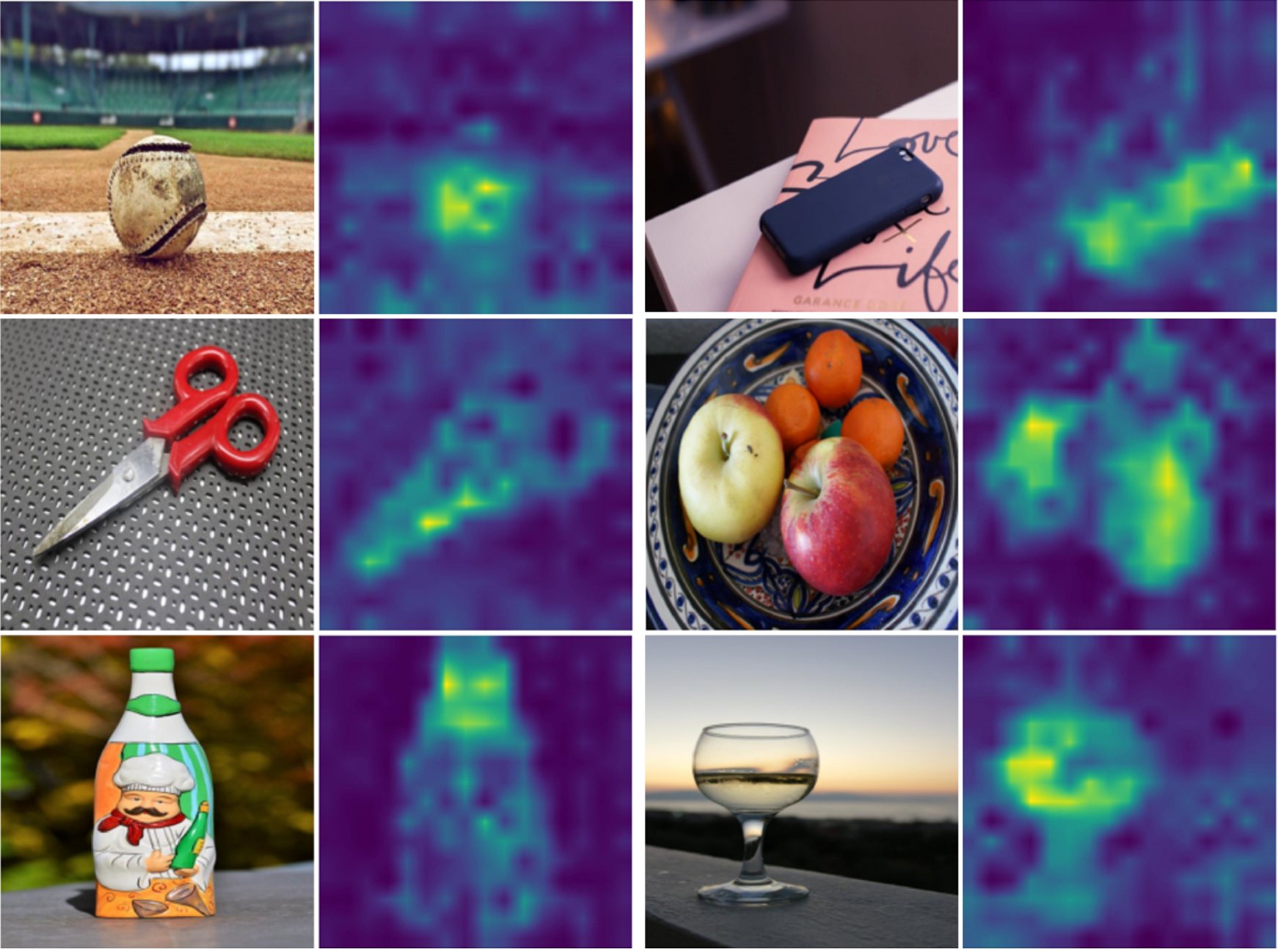}
   \caption{Visualization of CLS-guided mask.}
   \label{fig:cls}
\end{figure}

In \cref{fig:cls}, we display the visualization of the CLS-guided mask from the proposed CLS-guided transformer decoder.
It can be seen that the mask primarily concentrates on foreground objects, thus facilitating the cross-attention within salient regions.

\subsection{Visualization of Unseen Affordances}
\begin{figure}[t]
  \centering
   \includegraphics[width=1.0\linewidth]{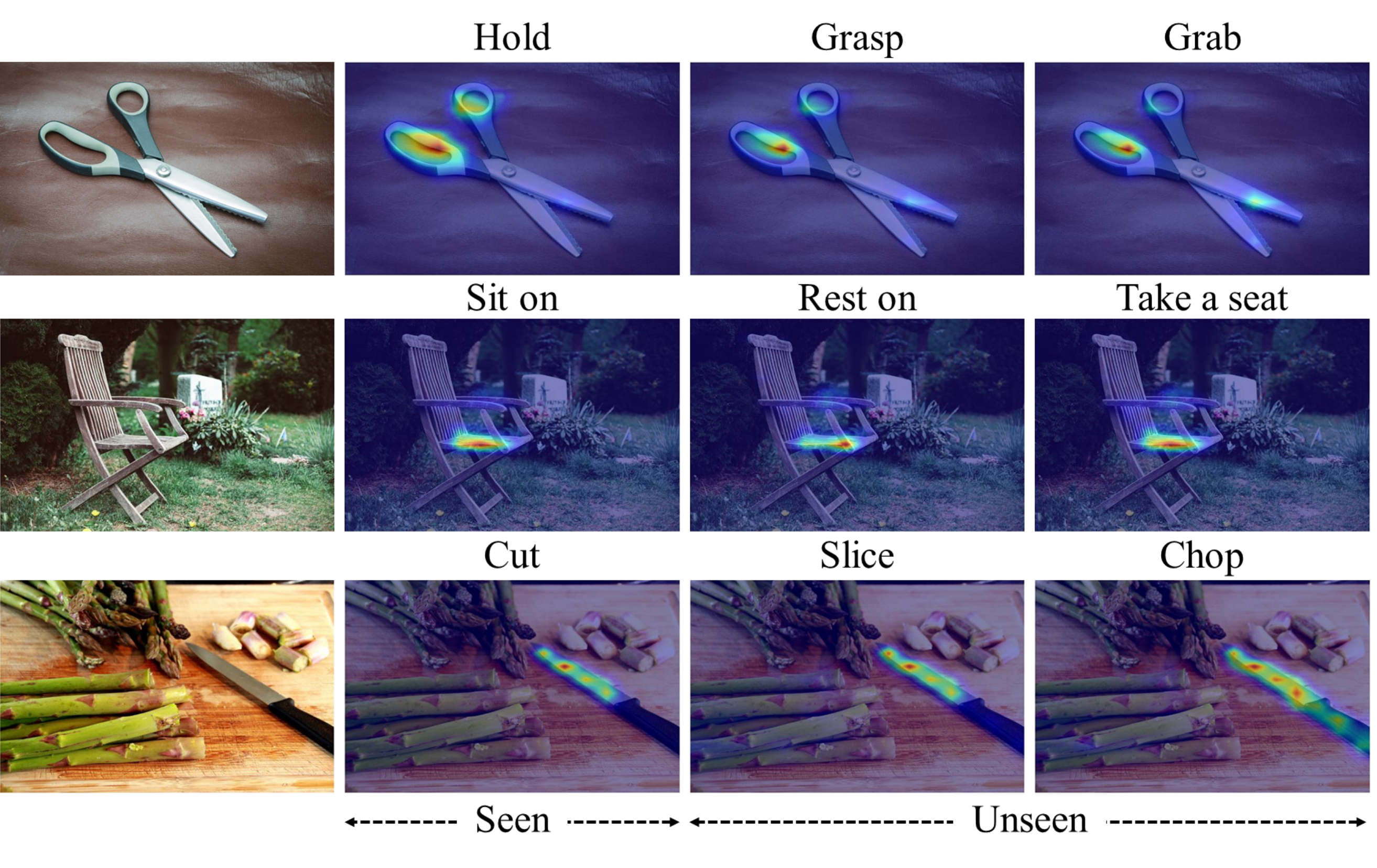}
   \caption{Qualitative examples of unseen affordance prediction on AGD20K dataset. The 2nd column shows the results on seen affordances, and the 3rd and 4th columns show results with unseen affordances.}
   \label{fig:novel_agd}
   \vspace{-3mm}
\end{figure}

In \cref{fig:novel_agd}, we further display examples on AGD20K dataset to showcase that our model has the ability to recognize unseen affordances.
It is evident that the model can consistently activate relevant affordance areas when receiving text that are previously unseen during training.

\subsection{Additional Qualitative Results}
\begin{figure*}[t]
  \centering
   \includegraphics[width=1.0\linewidth]{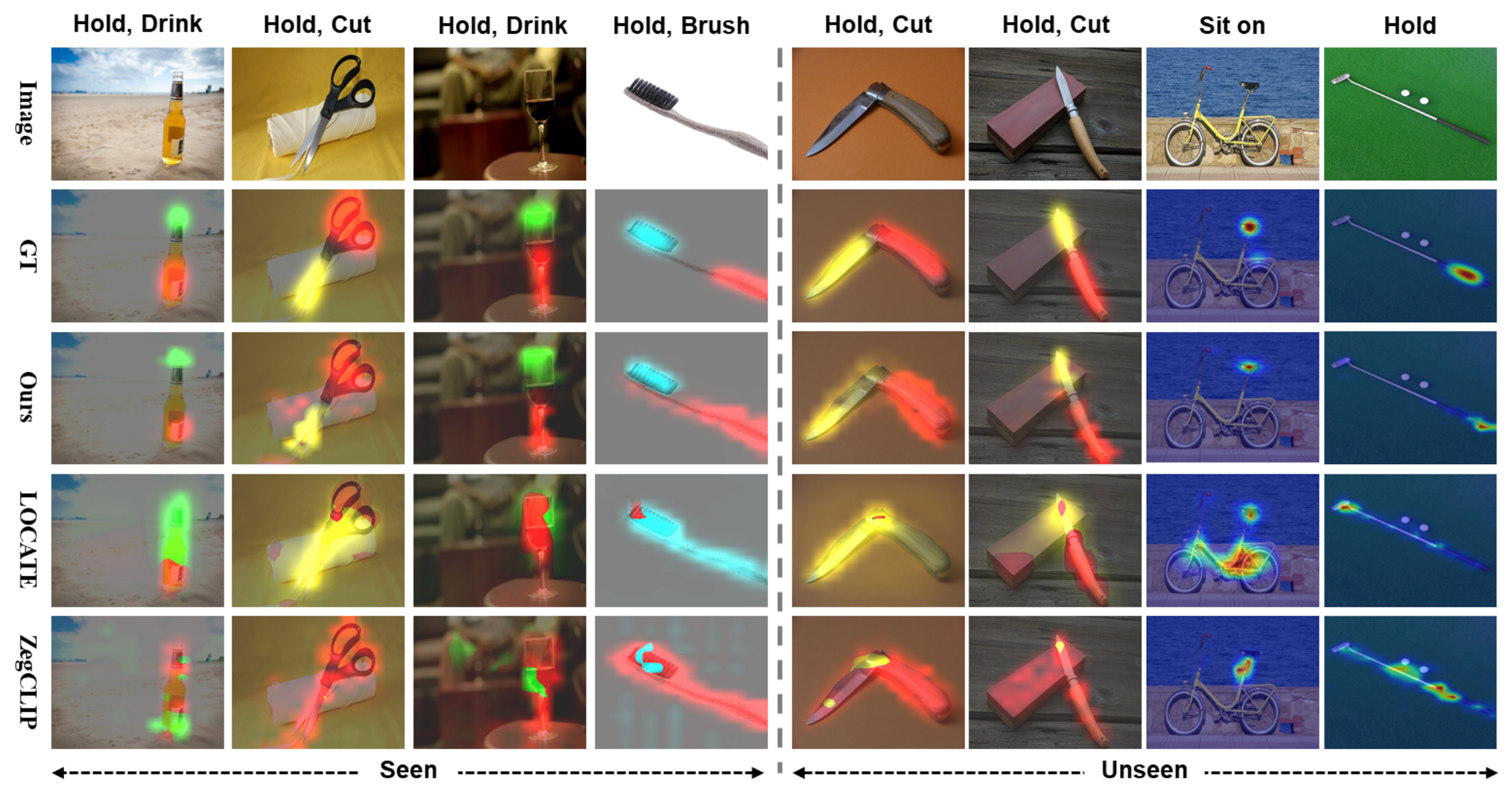}
   \caption{Additional qualitative comparison on AGD20K dataset.}
   \label{fig:supp_com_agd}
   \vspace{-3mm}
\end{figure*}

In \cref{fig:supp_com_agd}, we present more qualitative results on AGD20K dataset.
The comparison demonstrates that predictions from our methods exhibit clear separation among object parts, while predictions from other approaches often bias towards one part or the whole object.
In particular, our methods can locate very fine-grained affordance areas even for unseen objects, such as the saddle of a bicycle for ``sit on", and the handle of a golf club for ``hold".

\section{Discussion and Limitations}
This study introduces a novel problem of OOAL, and presents a framework built upon foundation models that can perform effective affordance learning with limited samples and annotations. 
We note that this framework can be potentially used in various applications, such as robotic manipulation and virtual reality.
For instance, in robotic manipulation, the model can make reasonable affordance predictions for diverse base and novel objects, requiring minimal annotation effort.
This stands in contrast to traditional methods that necessitate extensive training data or numerous simulated interaction trials to gain affordance knowledge.

Despite achieving good performance with few training samples, our framework reveals two limitations:
First, while text prompt learning enhances the performance within unseen objects, it diminishes the framework's generalization capacity to unseen affordances.
This occurs due to an excess of learnable tokens potentially weakening the intrinsic word similarities within the CLIP text encoder.
A viable solution to this limitation involves combining the learnable prompts with manually designed prompts.
Second, the performance is notably influenced by the selection of the one-shot example.
Instances with heavy occlusion or inferior lighting conditions can impact the learning performance.
Given the inherent challenges in learning from merely one-shot example, this limitation appears reasonable and logical.
\end{document}

%% file: 1_introduction.tex
\section{Introduction}
Affordances are the potential ``action possibilities" regions of an object~\cite{gibson, aff_vis_survey}, which play a pivotal role in various applications, including robotic learning~\cite{geng2022end, aff_detect_task_specific_grasp, aff-landscape}, scene understanding~\cite{scene-aff, learn2act_properly, learn_seg_aff}, and human-object interaction~\cite{hou2021affordance, grounded}.
In particular, affordance is crucial for embodied intelligence, since it facilitates agents' understanding of the associations between objects, actions, and effects in dynamic environments, thus bridging the gap between passive perception and active interaction~\cite{mon2007aff, cruz2016aff}.

Learning to recognize object affordances across a variety of scenarios is  challenging, since different objects can vary significantly in appearance, shape, and size, yet have the same functionality.
For instance, a chef's knife and a pair of office scissors share common affordances of cutting and holding, but their blades and handles look  different.

A large portion of the work~\cite{3d-affordancenet, nguyendetecting, affordancenet, learn2act_properly, aff_with_CNN_umd, bay-aff} has focused on learning a mapping between visual features and affordance labels, utilizing diverse resources as inputs, such as 2D images, RGB-D data, and 3D point clouds.
This mapping can be established through a labeled dataset with predefined objects and affordances.
However, large-scale affordance datasets are scarce, and most of them have a small number of object categories, making it difficult to apply the learned mapping to novel objects and scenes.
To reduce the reliance on costly annotation, some recent studies perform affordance learning from sparse key points~\cite{weakly_supervised_affordance_detection, strap_aff, weakaff2}, videos of humans in action~\cite{grounded, demo2vec, joint_hand_hotspot}, or human-object interaction images~\cite{ag_from_exocentric_imgs, locate}.
While alleviating the need for dense pixel labeling, these methods still require a large amount of training data. 
In addition, they often struggle to generalize to unseen objects and cannot identify novel affordances.

\begin{figure}[t]
  \centering
   \includegraphics[width=.95\linewidth]{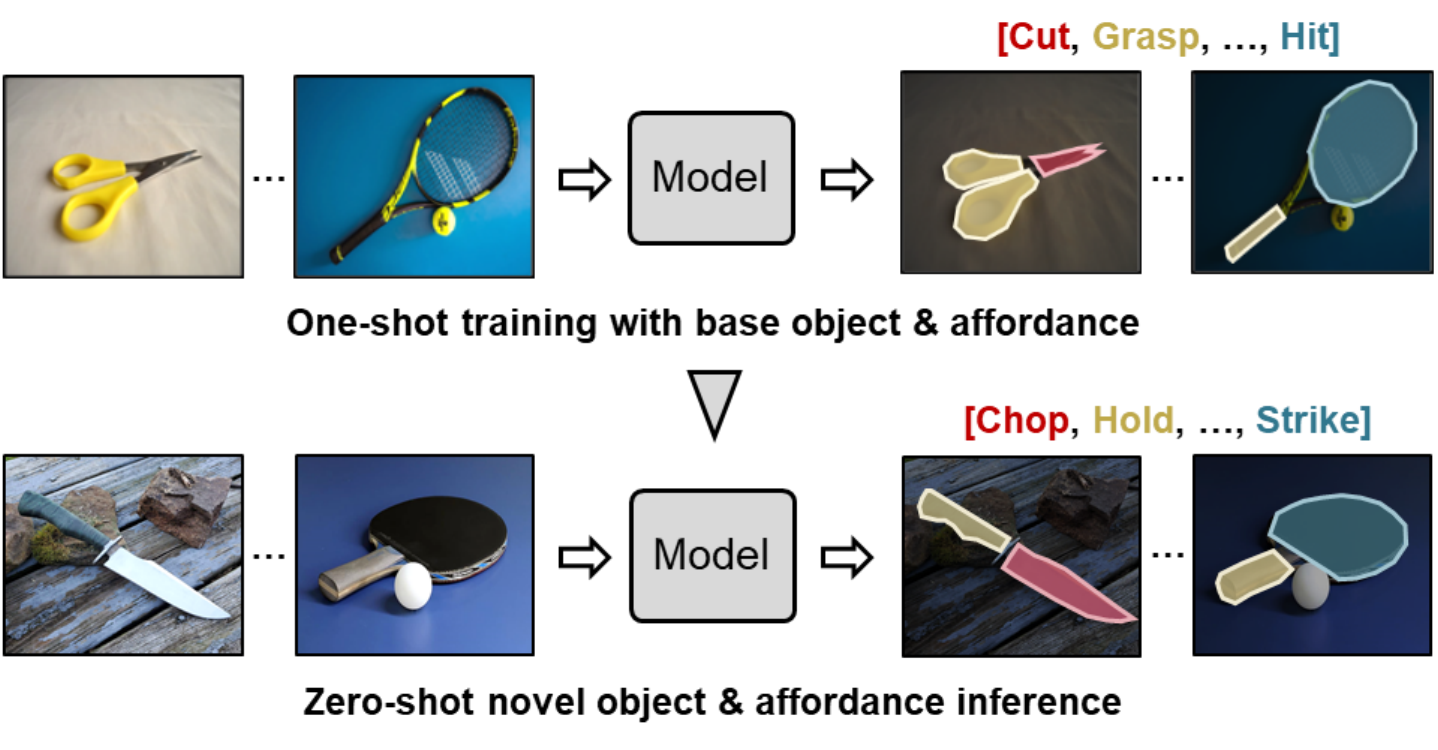}
   \caption{The pipeline of one-shot open affordance learning. It uses one image per base object for training, and can perform zero-shot inference on novel objects and affordances.}
   \label{fig:teaser}
\end{figure}

To tackle the above limitations, we are interested in learning an affordance model that does not rely on extensive datasets, and can comprehend novel object and affordance classes.
For example, after a model is trained with the knowledge that scissor blades afford cutting, it should generalize to related objects such as knives and axes, inferring that their blades can cut objects too.
Moreover, the model should be able to reason about semantically similar vocabularies,  \eg, ``hold" and ``grasp", ``cut" and ``slice", instead of knowing only predefined affordance categories. 

In this paper, we target the extreme case of using merely one example from each base object category and term this research problem as One-shot Open Affordance Learning (\p{}), where the model is trained with very little data, and is expected to recognize novel objects and affordances during inference.
The illustration of \p{} pipeline is shown in~\cref{fig:teaser}.
Compared with the typical affordance learning that requires numerous training samples and can only reason within a closed affordance vocabulary, \p{} alleviates the need of large-scale datasets and broadens the scope of inference.

To this end, we note that foundation Vision-Language Models (VLMs) can be a potential solution, which have recently emerged as powerful tools for a wide array of computer vision tasks.
The open vocabulary nature of these VLMs like CLIP~\cite{clip} that are trained on a large corpus of image-text data enables reasoning of previously unseen objects, scenes, and concepts.
However, we observe that these models often fail to understand nuanced vocabularies such as affordances or object parts. 
One hypothesis is that object parts and affordances appear much less frequently in image captions compared with objects.
Therefore, the following question naturally arises: 
\textit{Can we teach foundation models to comprehend more subtle, fine-grained aspects of objects, such as affordances, with very few examples?}
In this way, the generalization capability of foundation models can be inherited with minimum annotation effort.

To achieve this, we first conduct a thorough analysis of several representative foundation models.
The objective is to delve into their inherent understanding of affordances, and figure out what visual representation is suitable for data-limited affordance learning.
Based on the analysis, we then build a learning architecture and propose several methods, including text prompt learning, multi-layer feature fusion, and a CLS-token-guided transformer decoder, that can facilitate the alignment between visual representation and affordance text embeddings.
Lastly, we select a dense prediction task, affordance segmentation, for evaluation and comparison with a variety of state-of-the-art models,
where we find that our methods can achieve higher performance with less than 1\% of the complete training data.

Overall, our contributions can be summarized as follows:
(1) We introduce the problem of \p{}, aiming to develop a robust affordance model that can generalize to novel object and affordance categories without the need of massive training data.
(2) We conduct a comprehensive analysis on existing foundation models to explore their potential for \p{}.
Following the analysis, we build a learning architecture with vision-language foundation models, and design several methods to improve the alignment between visual features and affordance text labels.
(3) We implement extensive experiments with two datasets on affordance segmentation to demonstrate the effectiveness of our learning pipeline, and observe significant gains over baselines with strong generalization capability.

%% file: 2_related_work.tex
\section{Related Work}

\noindent\textbf{Affordance Learning.}
The term ``affordance'' is popularized by the psychologist James Gibson, who describes it as the properties of an object or the environment that suggest possible actions or interactions.
Building on this, researchers have developed many approaches to acquire affordance information in various ways.
In computer vision, initial research \cite{cad120, affordancenet, learn2act_properly, aff_obj_parts} has focused on affordance detection using convolutional neural networks.
As manual affordance annotations are often costly to acquire, much subsequent research has shifted its focus to weak supervision such as keypoints \cite{weakly_supervised_affordance_detection, weakaff2, strap_aff} or image-level labels \cite{ag_from_exocentric_imgs, grounded}.
Recent work has explored a novel perspective on how to ground affordances from human-object interaction images \cite{ag_from_exocentric_imgs, locate, 3d_aff_2d_img} or human action videos \cite{2023affvideo, demo2vec, grounded, joint_hand_hotspot}.
In robotics, affordance learning enables robots to interact effectively and intelligently with complex and dynamic environments \cite{aff_robot_survey, aff_rl_survey}.
Specifically, some work \cite{aff_detect_task_specific_grasp, aff_semantic_relations, aff_relations_2} utilizes affordance to build relationships between objects, tasks, and manipulations for robotic grasping. 
Other studies focus on learning affordance from other available resources that can be deployed on real robots, such as human teleoperated play data \cite{borja2022affordance}, image pairs \cite{bharadhwaj2023visual}, and egocentric video datasets \cite{bahl2023affordances}.

In contrast to the works above that often require a large amount of training data, we propose the problem of \p{} that aims to perform affordance learning with one  sample per base object category, and allows zero-shot inference to handle novel objects and affordances.

\vspace{1mm}
\noindent\textbf{Foundation Models for Affordance Learning.}
With the rapid development of foundation models such as Large Language Models (LLMs) and vision-language models, many research efforts have explored their utilization in affordance learning or reasoning.
Mees \etal~\cite{lang_aff_unstructured} leverage GPT-3~\cite{gpt3} to break down language instructions into subgoals, and learn a visual affordance model to complete real world long-horizon tasks.
Li \etal~\cite{locate} adopt DINO-ViT features to perform affordance grounding by transferring affordance knowledge from human-object interaction images to egocentric views.
Huang \etal~\cite{voxposer} propose a novel pipeline that uses LLMs~\cite{gpt4} for affordance reasoning, which interacts with VLMs to produce 3D affordance maps for robotic manipulation.
Recent studies \cite{lan-rf, lan-grasp, 6dof_part_aff} delve into the integration of affordance and language models for task-oriented grasping, which allows robots to grasp objects in a more appropriate and safe manner.

The closest methods to ours are AffCorrs~\cite{affcorrs} and OpenAD~\cite{ov-aff-3d}. 
AffCorrs utilizes the visual foundation model DINO to find corresponding affordances in a one-shot manner, but relevant objects are explicitly selected as support images to significantly reduce the difficulty.
OpenAD takes advantage of CLIP for open-vocabulary affordance detection in point clouds.
It requires a large number of manual annotations, while our work performs affordance learning with merely one example per base object category.

%% file: 3_proposed_method.tex
\section{Problem Setting}
One-shot Open Affordance Learning (\p{}) aims to learn a model to predict affordance with one example per base object class and can generalize to novel object classes.
In this work, we focus on the dense prediction task of affordance segmentation.
Specifically, objects are first divided into $N_b$ base classes and $N_o$ novel classes without intersection. 
The model receives only $N_b$ samples during training, one for each base object category, which is a pair of image $I \in\mathbb{R}^{H\times W \times 3}$ and pixel-wise affordance annotation $M \in\mathbb{R}^{H\times W \times N}$ ($N$ is the number of affordance categories in the dataset).
After training, evaluation is performed 
on the combination of base and novel object categories to measure the generalization ability of the model.
Also, affordance labels can be replaced with novel vocabularies that share similar semantics, such as ``chop", ``slice", and ``trim" to represent affordance akin to ``cut". 

It is worth noting that \p{} is different from one-shot semantic segmentation (OSSS)~\cite{one-sss} and one-shot affordance detection (OS-AD)~\cite{one-shot_aff_detect}.
Both OSSS and OS-AD receive one-shot sample during training. 
However, the sample keeps changing in each iteration, so the model can be exposed to many different images. %
Additionally, a support image is required at inference to provide prior information.
In comparison, \p{} performs one-shot training and zero-shot inference, which poses additional challenges.
The model needs to generalize to previously unseen objects, necessitating the ability to understand and recognize semantic relationships between seen and unseen classes with very limited data.

\section{Method}
\subsection{Analysis of Foundation Models}
\label{sec:analysis}
\begin{figure}[t]
  \centering
   \includegraphics[width=1.0\linewidth]{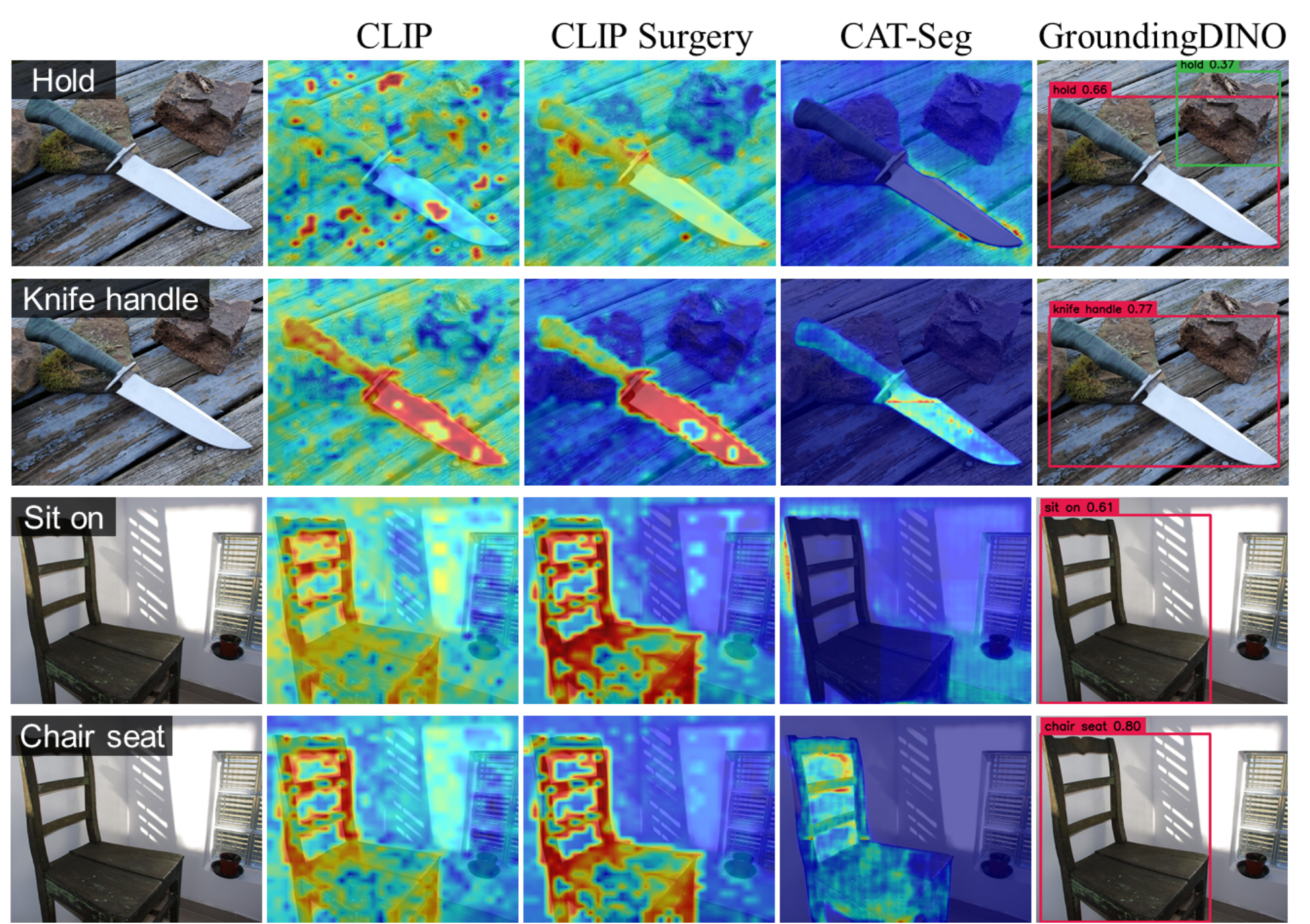}
   \caption{\textbf{Analysis of vision-language foundation models} on text-based affordance grounding. The 1st and 3rd rows use affordance texts as input queries, and the 2nd and 4th rows use corresponding object parts as input text queries.
   Visualizations show that these models have limited ability to recognize fine-grained affordances and object parts.
   }
   \label{fig:analysis-q1}
\end{figure}

\begin{figure*}[ht]
  \centering
   \includegraphics[width=0.9\linewidth]{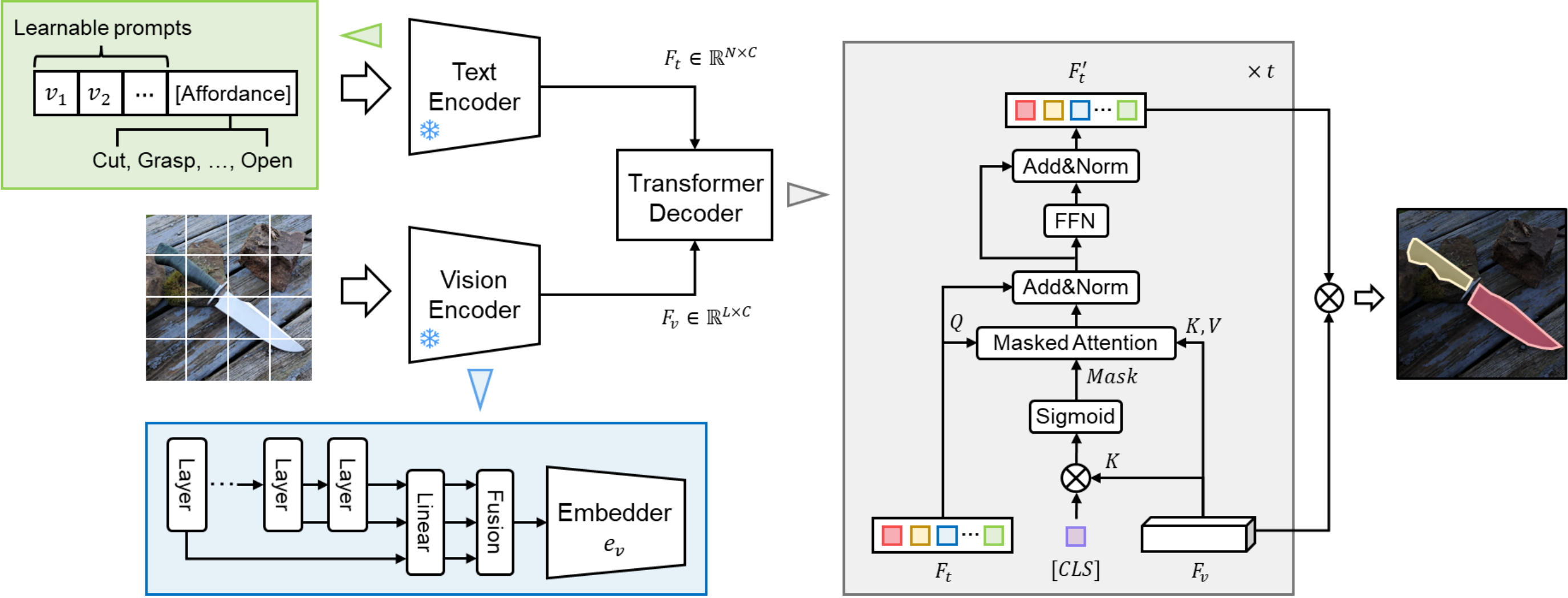}
   \caption{Proposed learning framework for \p{}. Our designs are highlighted in three color blocks, which are text prompt learning, multi-layer feature fusion, and CLS-guided transformer decoder. [CLS] denotes the CLS token of the vision encoder.}
   \label{fig:pipeline}
\end{figure*}

\begin{figure}[t]
  \centering
   \includegraphics[width=.9\linewidth]{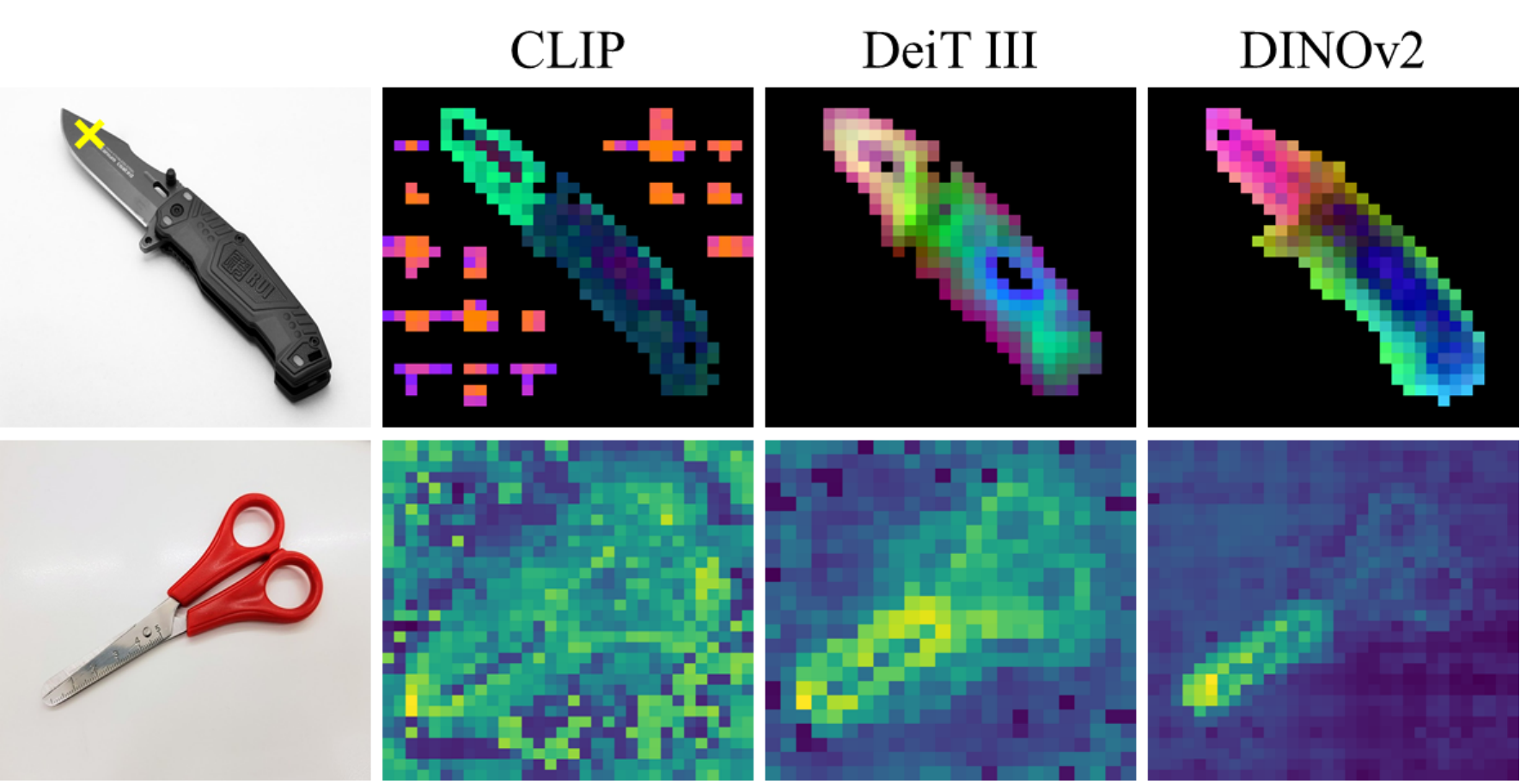}
   \caption{\textbf{Analysis of visual foundation models} on affordance learning. Top row: visualizations of PCA components. Bottom row: feature similarity maps between the yellow mark on the knife blade and the image of scissors. Qualitative results show that DINOv2 has clearer part-aware representations and better part-level semantic correspondence.}
   \label{fig:analysis-q2}
\end{figure}

The field of computer vision has recently witnessed a surge in the prevalence of large foundational models, such as CLIP \cite{clip}, Segment Anything \cite{sam}, DINO \cite{dino, dinov2} \etc.
These models exhibit strong zero-shot generalization capabilities for several computer vision tasks, making them seem like a great option to tackle the problem of \p{}.
To this end, we perform analysis on several existing foundation models which we split into three parts: 
~\ding{182} Do current vision-language foundation models and their variants have the ability to detect affordances via affordance/part-based prompting? ~\ding{183} Can the features of visual foundation models discriminate affordance regions in images?
and~\ding{184} Can these models generalize affordance recognition to novel objects and perform well in the low-shot setting?

Driven by question~\ding{182}, 
we select four representative models, \ie, the vanilla CLIP, a CLIP-based explainability method CLIP Surgery~\cite{clip_surgery}, a state-of-the-art open-vocabulary segmentation method CAT-Seg~\cite{catseg}, and an open-vocabulary detection method GroundingDINO~\cite{groundingdino}.
For vanilla CLIP, we employ the method proposed in MaskCLIP~\cite{maskclip} that directly extracts dense predictions without fine-tuning.
We use the text prompt template of ``somewhere to [\textit{affordance}]" to query visual features to find corresponding areas.
As illustrated in~\cref{fig:analysis-q1}, we note that most models cannot understand affordance well, except the detection model GroundingDINO, but its predictions mainly focus on the whole object rather than parts.
As for dense prediction models, CAT-Seg often recognizes affordance regions as background, and CLIP gives high activation on both foreground and background.
In comparison, CLIP Surgery fails to localize the ``holding" area for a knife, but manages to associate the phrase ``sit on" with a chair.
Furthermore, even when the affordance text is replaced with corresponding object parts, predictions from CLIP and GroundingDINO remain biased toward objects, while CLIP Surgery and CAT-Seg tend to activate the wrong parts.
This is consistent with recent findings \cite{ov-parts, denser_ovps} that CLIP has limited part recognition ability.

To answer questions \ding{183} and \ding{184}, 
we consider two essential characteristics of a good affordance model in the low-shot setting: (1) Part-aware representation. The visual representation should exhibit awareness of object parts, given that affordance often denotes small and fine-grained regions, \eg, a bicycle saddle to sit on or a knife handle to hold.
(2) Part-level semantic correspondence. This property is critical for generalization, since the model requires the understanding of semantic relations to make reasonable predictions on novel objects.
In addition, good correspondence proves advantages in scenarios with limited data, as the model can be more robust to intra-class recognition, and less susceptible to changes in appearance.
We then analyze the features from three representative and powerful visual foundation models, \ie, vision-language contrastive learning CLIP, fully-supervised learning DeiT III~\cite{deit3}, and self-supervised learning DINOv2.
First, we perform the principal component analysis (PCA) on the extracted patch features of each model to investigate the part awareness.
Visualization of PCA components in the top row of \cref{fig:analysis-q2} shows that all three models have part-aware features to some extent, yet CLIP cannot well distinguish the background, and features of DeiT III are not discriminative enough for different parts.
Next, we choose a different object that has equivalent affordances, \ie, knife and scissors, to assess the semantic correspondence.
The bottom row of \cref{fig:analysis-q2} shows the feature similarity maps computed as the cosine similarity between one patch representation on the knife blade and an image of scissors.
It is obvious that DINOv2 shows finer correspondence between blades of knife and scissors.
By contrast, CLIP produces messy correspondences in both foreground and background, and feature correspondences of DeiT III are only discriminative at the object level, but not specific to the affordance part region (cut).
From the above analysis, we conclude that DINOv2 is well suited for affordance learning due to its fine-grained part-aware representation and superior part-level semantic correspondence.
Quantitative comparisons are shown in \cref{sec:ablation}.

\subsection{Motivation and method}

Through a systematic analysis, we identify DINOv2 as a powerful tool for addressing the \p{} problem.
However, there are still fundamental issues that hinder performance in this challenging setting.
The first is that DINOv2 is a vision-only model, and lacks the ability to identify novel affordances.
One potential solution involves integrating a text encoder like CLIP, but it is recognized that the input text is sensitive to prompts.
This is particularly problematic in the case of affordances, which combine both an object and a verb, making manual prompt design a complex task.
The second issue is that while features of DINOv2 are part-oriented, the level of granularity varies across layers.
Determining the appropriate granularity level is crucial when handling affordances associated with diverse objects.
The third issue arises due to the absence of alignment between the DINOv2 vision encoder and CLIP text encoder, as they are trained separately and independently of each other.
Building upon these observations, we establish a vision-language framework based on DINOv2 and CLIP, and propose three modules to resolve each of the three fundamental bottlenecks mentioned above.

In this section, we first describe the overview of our proposed learning framework that builds on the powerful foundation models.
Then, we elaborate on the three proposed designs that help in the challenging \p{} problem.
Finally, we discuss the framework's capability to identify novel objects and affordances at inference.

\noindent \textbf{Overview.} The proposed learning framework is presented in \cref{fig:pipeline}, which consists of a vision encoder, a text encoder, and a transformer decoder.
First, the pretrained vision encoder DINOv2 is used to extract dense patch embeddings $\hat{F}_v \in\mathbb{R}^{L\times C_v}$, where $L$ is the number of tokens or patches.
Then, affordance labels are processed by the CLIP text encoder to obtain text embeddings $F_t \in\mathbb{R}^{N\times C}$.
To cope with inconsistent dimensions between visual and text embeddings, an embedder $e_v : \mathbb{R}^{C_v} \rightarrow \mathbb{R}^{C} $ with a single MLP layer is employed.
In the end, the lightweight transformer decoder takes both visual and text embeddings as input, and outputs the affordance prediction.

\noindent \textbf{Text Prompt Learning.}
Manually designing prompts for affordances can be a complicated work, especially considering that CLIP has difficulty in recognizing affordance (see~\cref{fig:analysis-q1}). 
Thus, we adopt the Context Optimization (CoOp)~\cite{coop} method to introduce automatic text prompt learning.
Instead of finetuning the CLIP text encoder, the inclusion of learnable prompts is an effective strategy that can alleviate the problem of overfitting and retain the inherent text recognition ability of CLIP.
Specifically, $p$ randomly initialized learnable context vectors $\{v_1, v_2, ..., v_p\}$ are inserted in front of the text CLS token, and they are shared for all affordance classes. 

\noindent \textbf{Multi-Layer Feature Fusion.}
Different layers of DINOv2 features often exhibit different levels of granularity \cite{deepvit}.
Since affordance may correspond to multiple parts of an object, a diverse set of granularities can be beneficial.
For this purpose, we aggregate the features of the last $j$ layers.
Each layer of features is first processed by a linear projection, and then all features are linearly combined with a weighted summation:
\begin{equation}
\hat{F}_v =  \sum_{i=1}^{j}\alpha_{i} \cdot  \phi(F_{n-i+1}), \quad
\alpha_1 + \alpha_2 + ... + \alpha_j = 1,
\end{equation}
where $F_n$ denotes the last layer, $\alpha$ is a learnable parameter that controls the fusion ratio of each layer, and $\phi$ indicates the linear transformation.
This straightforward fusion scheme enables adaptive selection among different granularity levels, allowing the model to handle affordance recognition across diverse scenarios.

\noindent \textbf{CLS-Guided Transformer Decoder.}
To deal with the lack of alignment between visual and text features, we propose a lightweight transformer decoder that applies a masked cross-attention mechanism to promote the mutual communication between two branches.
Since the [CLS] token of a foundation model is used in the computation of objective function, it often carries rich prior information of the whole image, such as salient objects or regions. 
Consequently, we utilize the [CLS] token to produce a guidance mask that constrains the cross-attention within a foreground region.

The decoder receives three inputs, \ie, text embeddings $F_t$, visual features $F_v$, and the [CLS] token $L_{cls}$.
Firstly, linear transformations are performed to yield query, key, and value:
\begin{equation}
Q=\phi_{q}(F_t), ~K=\phi_{k}(F_v), ~V=\phi_{v}(F_v).
\end{equation}
Here we use text embeddings as query, and visual features as key and value, allowing the model to focus on the update of text embeddings by retrieving relevant visual information that corresponds to the affordance text.
Next, the CLS-guided mask is calculated between the [CLS] token and key via matrix multiplication:
\begin{equation}
M_{cls} = \text{sigmoid}(\frac{\phi_c(L_{cls})K^T}{\sqrt{d_k}}),
\end{equation}
where $d_k$ is a scaling factor that equals the dimension of the keys.
The masked cross-attention is then computed as:
\begin{equation}
\hat{F}_t = \text{softmax}(QK^T/\sqrt{d_k})\cdot {M}_{cls}V + F_t.
\end{equation}
After that, the updated text embeddings $F_t^{'}$ are obtained by sending $\hat{F}_t$ through a feed-forward network (FFN) with a residual connection.
The decoder comprises $t$ layers of transformers, and the ultimate prediction is generated by performing matrix product between the output of the last transform layer and original visual features $F_v$, thereby ensuring the maximum retention of part-aware representations from DINOv2.
Lastly, binary cross entropy is employed as loss function to optimize parameters of linear layers, embedder, and decoder.

\noindent \textbf{Inference on novel objects and affordances.} 
During the training process, the decoder learns to establish an alignment between visual features and affordance text embeddings.
When encountering a novel object at inference, the aligned affordance text embeddings can locate corresponding object regions, leveraging the part-level semantic correspondence property inherent in DINOv2.
Similarly, as the model processes novel affordance text inputs, the generated text embeddings can also retrieve the aligned visual features, which are based on the semantic similarities to the base affordances seen in the training.

%% file: 4_experiments.tex
\section{Experiments}
\label{sec:exp}

\input{exp_tables}

\subsection{Datasets}
We choose two typical datasets, AGD20K~\cite{ag_from_exocentric_imgs} and UMD part affordance~\cite{aff_with_CNN_umd}, both of which include a large number of object categories that help in the evaluation of novel objects.
AGD20K is a large-scale affordance grounding dataset with 36 affordances and 50 objects, containing 23,816 images from exocentric and egocentric views.
It aims to learn affordance from human-object interaction images, and perform affordance localization on egocentric images.
As it is a dataset for weakly-supervised learning, images in the training set only have image-level labels.
Therefore we manually annotate 50 randomly selected egocentric images from each object category for training.
AGD20K also has two train-test splits for seen and unseen settings, and we follow their splits to evaluate the performance.
Note that AGD20K uses sparse annotation, where ground truth consists of keypoints within affordance areas, and then a gaussian kernel is applied over each point to produce dense annotation.

UMD dataset consists of 28,843 RGB-D images with 7 affordances and 17 object categories, and images of each object are captured on a revolving turntable.
It has two train-test splits termed category split and novel split.
We use the category split to evaluate base object categories and novel split to evaluate performance on novel object classes.
Due to its small number of object categories, we take one example from each base object instance to form the training set.
Specific affordance categories and object class splits can be found in the supplementary material.

\subsection{Implementation details}
Experiments are implemented on two GeForce RTX 3090 GPUs.
All visual foundation models use the same base-sized vision transformer (ViT-base).
We train the model using SGD optimizer with learning rate 0.01 for 20k iterations.
For experiments on AGD20K, images are first resized to 256 × 256 and randomly cropped to 224 × 224 with horizontal ﬂipping.
Experiments for UMD dataset are conducted on the opensource toolbox MMSegmentation~\cite{mmseg} with the default training setting.
The hyperparameters $p$, $j$, and $t$ are set to 8, 3, and 2, respectively.

Following previous work, we adopt the commonly used Kullback-Leibler Divergence (KLD), Similarity (SIM), and Normalized Scanpath Saliency (NSS) metrics to evaluate the results on AGD20K.
For UMD dataset, we use the metric of mean intersection-over-union (mIoU), and also incorporate the harmonic mIoU as a balanced measure that accounts for both seen and unseen settings.

\begin{figure*}[t]
  \centering
   \includegraphics[width=.95\linewidth]{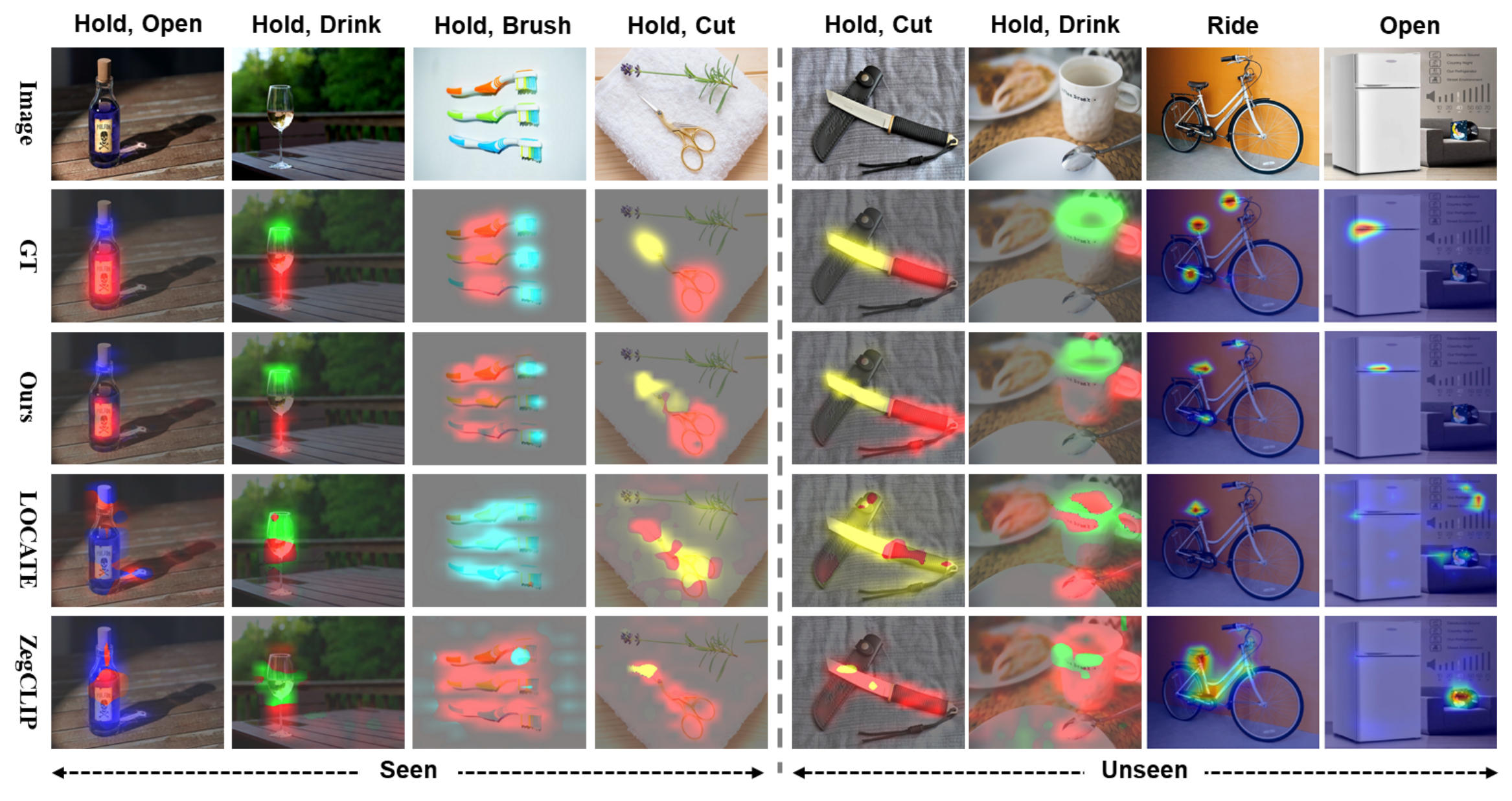}
   \caption{Qualitative comparison with LOCATE and ZegCLIP on AGD20K dataset. When multiple affordance predictions overlap, the one with higher value is displayed. Our predictions distinguish different object parts, while other methods often make overlapping predictions.}
   \label{fig:agd_com}
\end{figure*}

\begin{figure}[t]
  \centering
   \includegraphics[width=1.0\linewidth]{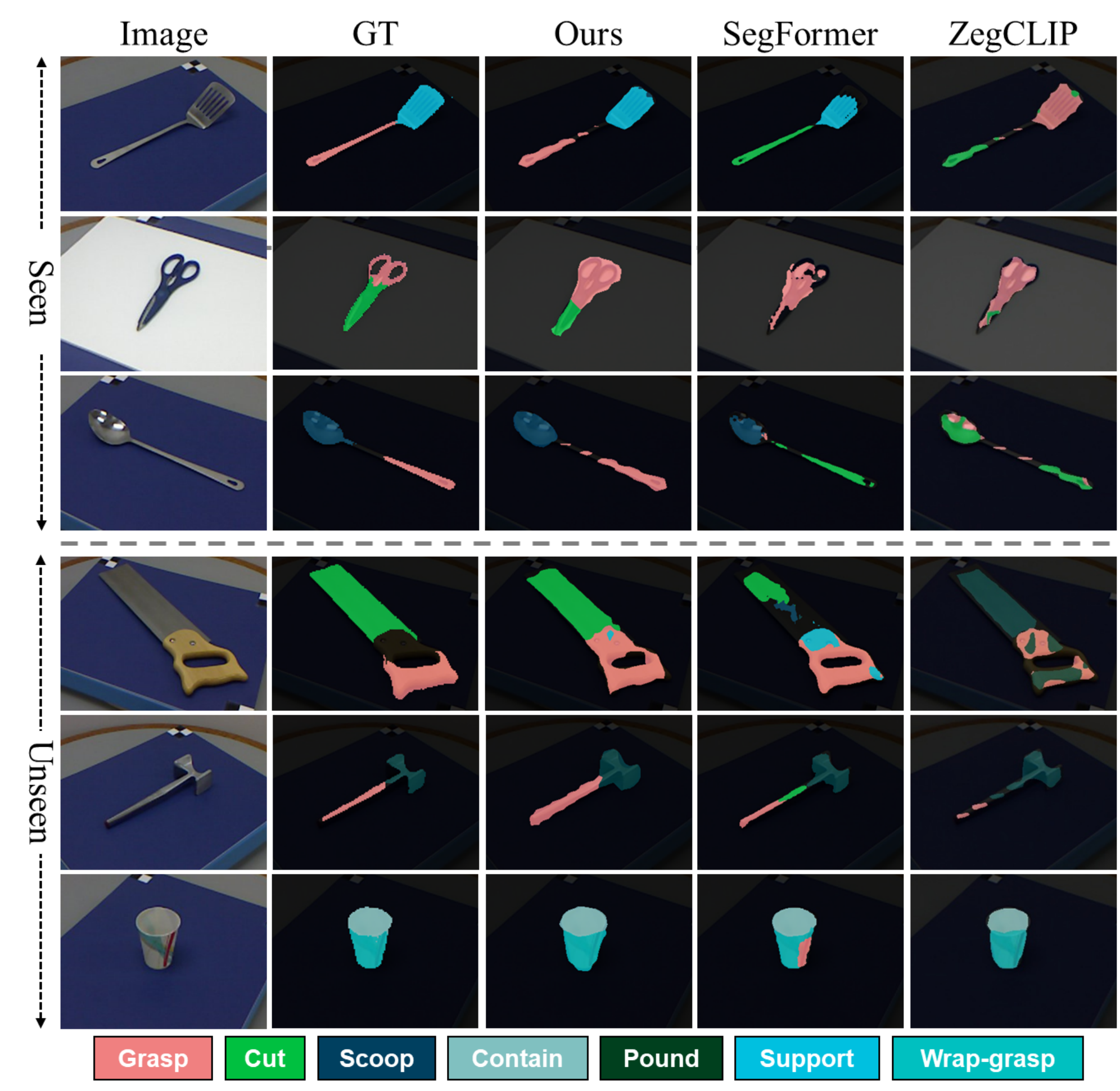}
   \caption{Qualitative comparison with SegFormer and ZegCLIP on UMD affordance dataset in \p{} setting. Images have been enlarged and cropped for better visualization.}
   \vspace{-3mm}
   \label{fig:umd_com}
\end{figure}

\subsection{Comparison to state-of-the-art methods}
AGD20K dataset is benchmarked with weakly supervised affordance grounding (WSAG) approaches, which use image-level object and affordance labels to do affordance segmentation.
Note that results from WSAG methods are not directly comparable to our setting, as training labels are different.
Despite using only image-level labels, the training data required are more than 460 times of ours.
The results in \cref{tab:com_agd} demonstrate that our results exceed all WSAG counterparts in an easy and realistic setting.
We also benchmark open-vocabulary segmentation methods of MaskCLIP, SAN, and ZegCLIP for further comparison.
We find that these CLIP-based methods have a large performance gap with ours, and are also inferior to the state-of-the-art WSAG method LOCATE.

The comprehensive comparison on UMD dataset is displayed in \cref{tab:com_umd}, where we benchmark the results with several representative semantic segmentation methods (PSPNet, DeepLabV3+, SegFormer) and open-vocabulary semantic segmentation methods.
For fair comparison, the classical segmentation methods are trained with the full training set, while foundation-model-based methods like ZegCLIP and SAN are evaluated in the \p{} setting.
It is clear that our proposed model is quite effective, which can be comparable to fully-supervised methods with only 0.36\% of their training data.
To explore how fully-supervised methods are affected by the limited data, we further train these models in the \p{} setting.
Results in \cref{tab:com_umd} show that the performance of these models degrades by around 10\% in both seen and unseen settings when given only one-shot example.
Additionally, under the same \p{} setting, we observe a more apparent gain over other CLIP-based open-vocabulary segmentation methods, showing that CLIP is not suitable for data-limited affordance learning.
The poor performance of MaskCLIP from both tables also verifies that CLIP has very limited understanding on affordance.

\begin{figure}[t]
  \centering
   \includegraphics[width=0.95\linewidth]{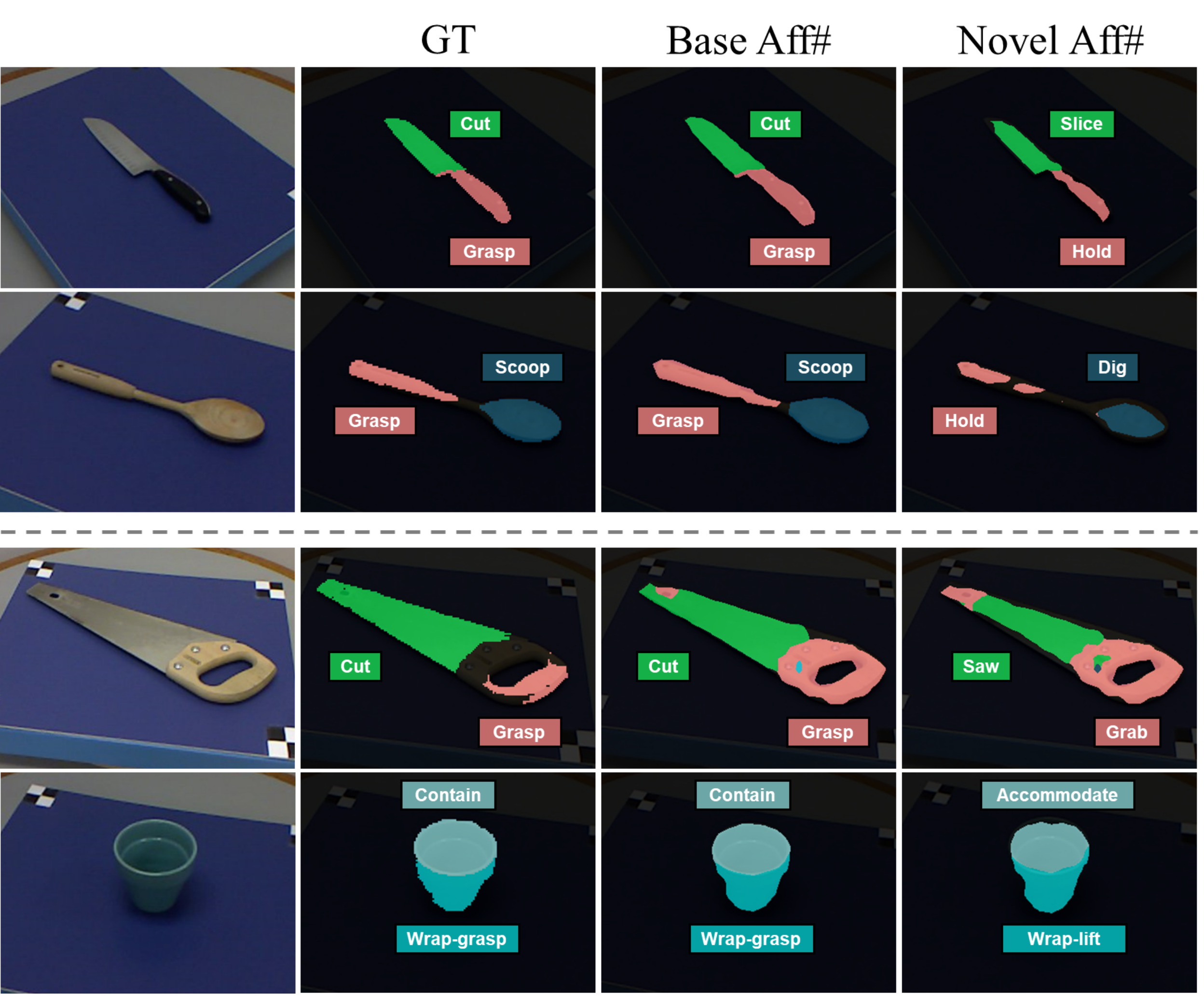}
   \caption{Qualitative examples of novel affordance prediction on UMD dataset. The 1st and 2nd rows display results on base objects, and the 3rd and 4th rows show results for novel objects.}
   \label{fig:novel_aff}
\end{figure}

\subsection{Qualitative results}
Qualitative comparisons on AGD20K dataset are shown in \cref{fig:agd_com}.
We note that WSAG methods like LOCATE often make overlapping predictions for examples with multiple affordances, while our results show a clear separation between different affordance regions.
ZegCLIP can make reasonable predictions to some extent, but it mostly focuses on the whole object and the accuracy is far from satisfactory, whereas our results are more part-focused, especially for the unseen objects.
For example, the prediction for the unseen object of bicycle show that our model can handle the complex affordance (ride) with multiple separated affordance areas (saddle, handlebar, and pedal).
In \cref{fig:umd_com}, we display the results for UMD dataset.
We observe that SegFormer and ZegCLIP often fail to recognize affordances of objects whose parts are similar in appearance.
Also, they tend to misclassify metallic object parts as cuttable affordance, suggesting that inferring affordances with only appearance features can be misleading.
In comparison, our predictions are more accurate due to the utilization of DINOv2's part-level semantic correspondences.

One particular feature of our model is that it can recognize novel affordances not shown during training.
To demonstrate this, we replace the original affordance labels with semantically similar words and check if the model can still reason about corresponding affordance areas.
As shown in~\cref{fig:novel_aff}, the model manages to make correct predictions for novel affordances, such as ``hold and grab" for base affordance ``grasp", ``saw" for ``cut", and ``accommodate" for ``contain".

\subsection{Ablation study}
\label{sec:ablation}
The ablation study is performed on the more challenging AGD20K dataset due to its natural images with diverse backgrounds.
Ablations on hyperparameters are left in the supplementary material.

\noindent\textbf{Different Vision Encoders.}
To complement the qualitative analysis in \cref{sec:analysis}, we conduct quantitative experiments on CLIP, DeiT III, and DINOv2.
Specifically, we simply process the visual features with the embedder, and perform matrix multiplication with pre-computed affordance text embeddings to output segmentation maps.
As shown in \cref{tab:abla1}, CLIP and DeiT III exhibit comparable performance, whereas DINOv2 achieves much better results in both seen and unseen settings, which are consistent with the analysis that DINOv2 is more suitable for affordance learning. 

\noindent\textbf{Proposed Methods.}
We use the DINOv2 with a simple embedder as baseline, and gradually integrate our methods to analyze the effect of each proposed design.
The results in \cref{tab:abla2} reveal that each module can consistently deliver notable improvements.
In particular, we notice that the inclusion of a transformer decoder can enhance the performance in the seen setting, but yield inferior results for the unseen setting.
With the integration of the CLS-guided mask, results of both settings can be improved, suggesting that restricting the cross-attention space is an effective strategy for unseen object affordance recognition.

\input{ablation_tables}

%% file: exp_tables.tex
\begin{table*}[!htb]
\centering
\small
\begin{tabular}{cclcccccc}
\toprule
\multirow{2.5}{*}{\textbf{Task}} & \multirow{2.5}{*}{\textbf{\makecell{Training Data \\ seen / unseen split}}} & \multirow{2.5}{*}{\textbf{Method}} &  \multicolumn{3}{c}{\textbf{Seen}} & \multicolumn{3}{c}{\textbf{Unseen}} \\ \cmidrule(lr){4-6} \cmidrule(lr){7-9}
&            &     & KLD$\downarrow$    & SIM$\uparrow$    & NSS$\uparrow$  & KLD$\downarrow$     & SIM$\uparrow$     & NSS$\uparrow$    \\ \midrule

\multirow{4}{*}{\makecell{WSAG}}  & 
\multirow{4}{*}{\makecell{23,083 / 15,543 images\\image-level labels}}
&  Hotspots \cite{grounded}                & 1.773  & 0.278  & 0.615 & 1.994   & 0.237   & 0.577  \\
                                                     &        &                                Cross-view-AG \cite{ag_from_exocentric_imgs}           & 1.538  & 0.334  & 0.927 & 1.787   & 0.285   & 0.829  \\
                                                                                        &       & Cross-view-AG+ \cite{ag_from_exocentric_imgs+}          & 1.489  & 0.342  & 0.981 & 1.765  & 0.279   & 0.882  \\
                                                                                               & &LOCATE~\cite{locate}                    & \underline{1.226} & \underline{0.401}  & \underline{1.177} & \underline{1.405}   & \underline{0.372}   & \underline{1.157} \\ \midrule
\multirow{4}{*}{\makecell{\p{}}}  & 
\multirow{4}{*}{\makecell{50 / 33 images\\keypoint labels}}
&  MaskCLIP~\cite{maskclip} & 5.752  & 0.169 & 0.041 &  6.052 & 0.152   & 0.047  \\
 &   &  SAN~\cite{san}  & 1.435 & 0.357 & 0.941 & 1.580   & 0.351  & 1.022 \\
 &   &  ZegCLIP~\cite{zegclip}  & 1.413 & 0.387  & 1.001 &  1.552  & 0.361  & 1.042 \\
& &Ours   &   \textbf{0.740}     & \textbf{0.577}       &    \textbf{1.745}    &  \textbf{1.070}       &   \textbf{0.461}     & \textbf{1.503} \\
\bottomrule
\end{tabular}
\caption{Comparison with state of the art on AGD20K dataset. \p{} setting uses 0.22\% / 0.21\% of the full training data. WSAG denotes weakly-supervised affordance grounding. The \textbf{best} and \underline{second-best} results are highlighted in bold and underlined, respectively.}
\label{tab:com_agd}
\end{table*}

\begin{table}[!htb]
\centering
\small
\begin{tabular}{clccc}
\toprule
\multicolumn{1}{c}{\textbf{Setting}}                                                                           & \textbf{Method}      & \textbf{Seen} & \textbf{Unseen} & \textbf{hIoU} \\ \midrule
\multicolumn{1}{c}{\multirow{3}{*}{\makecell{Fully\\Supervised}}}  & 

DeepLabV3+~\cite{deeplabv3+}  & \multicolumn{1}{c}{70.5}         & \multicolumn{1}{c}{57.5}         & 63.3    \\
 & SegFormer~\cite{segformer} & \multicolumn{1}{c}{\underline{74.6}}         & \multicolumn{1}{c}{57.7}         & 65.0     \\ 
&PSPNet~\cite{pspnet}      & \multicolumn{1}{c}{72.0}         & \multicolumn{1}{c}{\textbf{60.8}}         & \underline{66.0}     \\ \midrule

\multirow{7}{*}{OOAL}     & PSPNet~\cite{pspnet}      &  56.7   &     46.6    & 51.1                        \\
 & DeepLabV3+~\cite{deeplabv3+}  &     56.8                         &     48.4                         &    52.3                      \\
& SegFormer~\cite{mask2former} & 64.6                             &   51.4                           &      57.3                    \\ \cmidrule(l){2-5}
                                                                              & MaskCLIP~\cite{maskclip}    & 4.25                             & 4.24                             & 4.25                         \\
                                                       & SAN~\cite{san}  & 45.1    &   32.2                           &    37.5                      \\
                                  & ZegCLIP~\cite{zegclip}     &  47.4                            &   36.0                           &     40.9                     \\
                                                      &                                                                                                      Ours        &                     \textbf{74.6}         &   \underline{59.7}                           &  \textbf{66.4}              \\ \bottomrule        
\end{tabular}
\caption{Comparison on UMD dataset. Fully-supervised methods are trained with 14,823 and 20,874 images with pixel-level labels for seen and unseen split, respectively. In contrast, \p{} setting uses 54 and 76 images, 0.36\% of the full training data.}
\label{tab:com_umd}
\end{table}

%% file: ablation_tables.tex
\begin{table}[!t]
\centering
\small
\begin{tabular}{@{}lcccccc@{}}
\toprule
\multirow{2.5}{*}{Model} & \multicolumn{3}{c}{Seen} & \multicolumn{3}{c}{Unseen} \\ \cmidrule(lr){2-4} \cmidrule(l){5-7} 
                        & KLD$\downarrow$    & SIM$\uparrow$    & NSS$\uparrow$    & KLD$\downarrow$     & SIM$\uparrow$     & NSS$\uparrow$    \\ \midrule
CLIP                     &   1.294     &     0.384   & 1.107       &     1.556    &    0.327     &  0.966      \\
DeiT III                  &      1.301  &  0.378      & 1.140       &   1.535      &  0.321       & 1.049    \\
DINOv2               &   1.156     & 0.425       &   1.297     &    1.462     &   0.360      &   1.105   
\\ \bottomrule
\end{tabular}
\caption{Ablation results of different visual foundation models.}
\label{tab:abla1}
\end{table}

\begin{table}[!t]
\centering
\small
\begin{tabular}{@{}lcccccc@{}}
\toprule
\multirow{2.5}{*}{Method} & \multicolumn{3}{c}{Seen} & \multicolumn{3}{c}{Unseen} \\ \cmidrule(lr){2-4} \cmidrule(l){5-7} 
                        & KLD$\downarrow$    & SIM$\uparrow$    & NSS$\uparrow$    & KLD$\downarrow$     & SIM$\uparrow$     & NSS$\uparrow$    \\ \midrule
Baseline                     &   1.156     & 0.425       &   1.297     &    1.462     &   0.360      &   1.105     \\ 
+ TPL                   &      1.060  &  0.455      & 1.422       &   1.338      &  0.390       & 1.302    \\
+ MLFF               &     0.846   & 0.537       &       1.622 &     1.115    &  0.447       & 1.440       \\
+ TD               &     0.749   & 0.578       &       1.738 &     1.131    &  0.443       & 1.408       \\
+ CTM                &   0.740     & 0.577       &    1.745    &  1.070       &    0.461     & 1.503
\\ \bottomrule
\end{tabular}
\caption{Ablation results of proposed modules. TPL: text prompt learning. MLFF: multi-layer feature fusion. TD: transformer decoder. CTM: CLS-guided mask.}
\vspace{-3mm}
\label{tab:abla2}
\end{table}

%% file: 5_conclusion.tex
\section{Conclusion}
In this paper, we propose the problem of one-shot open affordance learning that uses one example per base object category as training data, and has the ability to recognize novel objects and affordances.
We first present a detailed analysis into different foundation models for the purpose of data-limited affordance learning.
Motivated by the analysis, we build a vision-language learning framework with several proposed designs that better utilize the visual features and promote the alignment with text embeddings.
Experiment results demonstrate that we achieve comparable performance over several fully-supervised baselines with less than 1\% of the full training data.

%% file: dataset_tab.tex
\begin{table*}[ht]
\centering
\small
\begin{tabular}{@{}cll@{}}
\toprule
Dataset & Affordance                                                                                                                                                                                                                                                            & Object                                                                                                                                                                                                                                                                                                                                                                                                                                                                               \\ \midrule
UMD     & \makecell[l]{(7) grasp, cut, scoop, contain,\\ pound, support, wrap-grasp}                                                                                           & \makecell[l]{(17) bowl, cup, hammer, knife, ladle, mallet,\\ mug, pot, saw, scissors, scoop, shears, shovel,\\ spoon, tenderizer, trowel, turner}                                                    \\ \midrule
AGD20K  & \makecell[l]{(37) beat, boxing, brush with, carry, catch, cut, cut with,\\ drag, drink with, eat, hit, hold, jump, kick, lie on, lift,\\ look out, open, pack, peel, pick up, pour, push, ride,\\ sip, sit on, stick, stir, swing, take photo, talk on,\\ text on, throw, type on, wash, write} & \makecell[l]{(50) apple, axe, badminton racket, banana, baseball, baseball bat,\\ basketball, bed, bench, bicycle, binoculars, book, bottle, bowl,\\ broccoli, camera, carrot, cell phone, chair, couch, cup,\\ discus, drum, fork, frisbee, golf clubs, hammer, hot dog,\\ javelin, keyboard, knife, laptop, microwave, motorcycle,\\ orange, oven, pen, punching bag, refrigerator, rugby ball,\\ scissors, skateboard, skis, snowboard, soccer ball, suitcase,\\ surfboard, tennis racket, toothbrush, wine glass} \\ \bottomrule
\end{tabular}
\caption{Affordance and object classes in the UMD and AGD20K dataset. The number of classes is shown in parentheses.}
\label{class_dataset}
\end{table*}

\begin{table*}[ht]
\centering
\small
\begin{tabular}{cll}
\toprule
Dataset & Base Objects (Train)                                                                                                                                                                                                                                                                                                                               & Novel Objects (Test)                                                                                                                     \\ \midrule
UMD     & \makecell[l]{(8) bowl, hammer, knife, mallet, mug, scissors, spoon, turner}     & \makecell[l]{(9) cup, ladle, pot, saw, scoop, shears,\\ shovel, tenderizer, trowel}                 \\ \midrule
AGD20K  & \makecell[l]{(33) apple, badminton racket, baseball, baseball bat, bench,\\ book, bottle, bowl, carrot, cell phone, chair, couch,\\ discus, fork, frisbee, hammer, hot dog, javelin,\\ keyboard, microwave, motorcycle, orange, oven,\\ punching bag, rugby ball, scissors, skateboard,\\ snowboard, suitcase, surfboard, tennis racket,\\ toothbrush, wine glass} & \makecell[l]{(14) axe, banana, basketball, bed, bicycle,\\ broccoli, camera, cup, golf clubs, knife,\\ laptop, refrigerator, skis, soccer ball} \\ \bottomrule
\end{tabular}
\caption{Object category division in the unseen split of UMD and AGD20K dataset. The number of categories is shown in parentheses.}
\label{unseen_dataset}
\end{table*}